\documentclass[10pt,twocolumn,letterpaper]{article}

\usepackage{iccv}
\usepackage{times}
\usepackage{epsfig}
\usepackage{graphicx}
\usepackage{amsmath}
\usepackage{amssymb}

\usepackage[breaklinks=true,bookmarks=false]{hyperref}
\usepackage{subfiles}

\iccvfinalcopy 

\def\iccvPaperID{10224} 

\ificcvfinal\fi

\title{An Image is Worth 16x16 Words, What is a Video Worth?}
\author{Gilad Sharir \hspace{0.1cm} Asaf Noy \hspace{0.1cm} Lihi Zelnik-Manor \\
DAMO Academy, Alibaba Group
}
\date{February 2021}

\begin{document}

\maketitle


\begin{abstract}
\noindent Leading methods in the domain of action recognition try to distill information from both the spatial and temporal dimensions of an input video. Methods that reach State of the Art (SotA) accuracy, usually make use of 3D convolution layers as a way to abstract the temporal information from video frames. The use of such convolutions requires sampling short clips from the input video, where each clip is a collection of closely sampled frames. 
Since each short clip covers a small fraction of an input video, multiple clips are sampled at inference in order to cover the whole temporal length of the video. This leads to increased computational load and is impractical for real-world applications. We address the computational bottleneck by significantly reducing the number of frames required for inference.
Our approach relies on a temporal transformer that applies global attention over video frames, and thus better exploits the salient information in each frame.  
Therefore our approach is very input efficient, and can achieve SotA results (on Kinetics dataset) with a fraction of the data (frames per video), computation and latency. Specifically on Kinetics-400, we reach $80.5$ top-1 accuracy with $\times 30$ less frames per video, and $\times 40$ faster inference than the current leading method.\footnote{Code is available at: https://github.com/Alibaba-MIIL/STAM}

\end{abstract}

\section{Introduction}
\begin{figure} [t]
  \centering
  \includegraphics[width=\columnwidth]{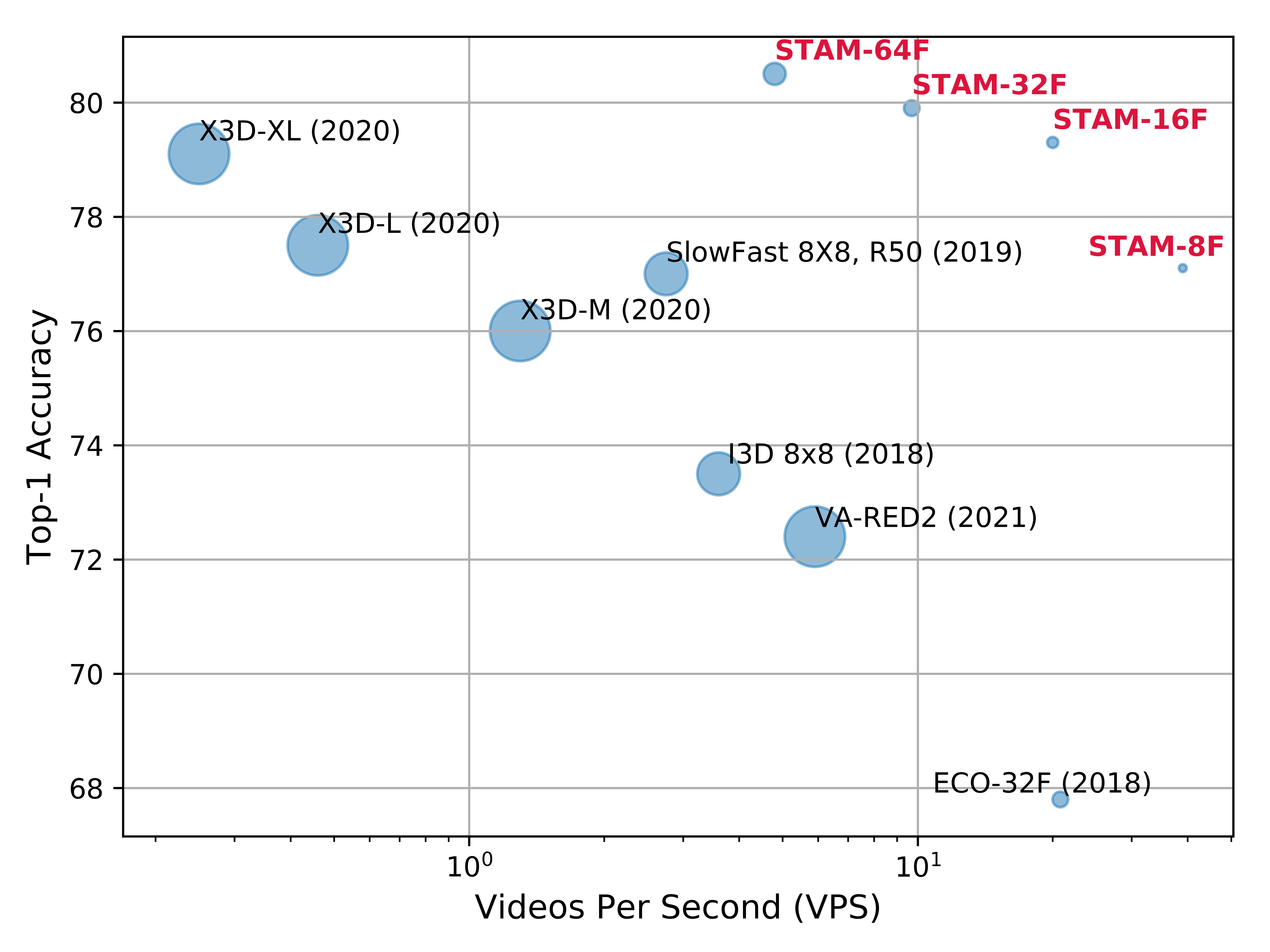}
  \caption{{\textbf{Kinetics-}$\mathbf{400}$ top-$1$ Accuracy vs Runtime, measured over Nvidia V100 GPU and presented in log-scale. Markers sizes are proportional to the number of frames used per video by leading methods.
  Our method provides dominating trade-off for those three properties.}}
  \label{fig:runtime_accuracy}
\end{figure} 

\begin{figure*} [t]
  \centering
  \includegraphics[width=2\columnwidth]{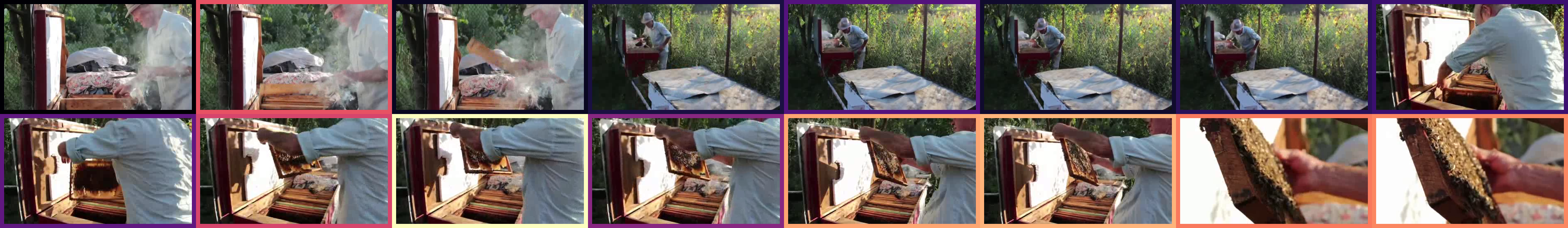}
  \caption{\textbf{Frame attention} 16 frames uniformly sampled from a 10 second input video depicting 'beekeeping'.  These frames are used as input to our model.  Each frame's border displays the attention weight of that frame corresponding to the classification token (in heatmap range).  We see that more attention is given to frames in which the action can clearly be identified. }
  \label{fig:frames_attn}
\end{figure*} 
\noindent 
The stellar growth in video content urges the need for more efficient video recognition.
Increased camera coverage and constantly growing network bandwidth for video streaming are making online recognition~\cite{liu2018ssnet,zolfaghari2018eco} essential in varied domains such as robotics, security and human-computer interaction. 
Additional applications like large-scale video retrieval benefit directly from faster recognition~\cite{araujo2017large}, as well as from efficient utilization of video frames transcoding. 

In action recognition the task is to classify a video 
by extracting relevant information from its individual frames.  
The success of Convolutional Neural Networks (CNN) over images has been utilized for action recognition via 3D convolutions, extracting both spatial and temporal information out of consecutive frames. 
Since 3D convolutions are computationally expensive,
the common practice is to apply those on a predefined number of short video clips, each composed of densely sampled frames, and average the predictions over these clips. 
Since the clips should cover the entire video for accurate predictions, a large fraction of the video frames is used by such methods, leading to computational bottlenecks of frames processing and transcoding. 
Recent methods addressed the processing bottleneck from different angles:
more efficient per-frame architectures~\cite{feichtenhofer2020x3d,ullah2021efficient} and 3D modules~\cite{tran2018closer,carreira2017quo}, clip sampling~\cite{zheng2020dynamic} and two-stream networks~\cite{zheng2020dynamic,simonyan2014two}.
While the trade-off between accuracy and efficiency is continuously improving, 
for many real-time applications the required runtime is lower by orders of magnitude than the ones offered by current state-of-the-art methods.

In this work, we take a different path towards efficient action recognition.
We train classifiers to learn spatio-temporal representations from small numbers of uniformly sampled video frames via an end-to-end attention mechanism. 
Our approach is motivated by human action recognition that is shown to maintain a similar accuracy when a small numbers of frames is viewed instead of the entire video~\cite{schindler2008action}.
Conversely, multiple clips inference with 3D convolutions is often involved with redundant computations as consecutive video frames tend to be similar. In addition, its scope is limited by design to short actions, while real-world applications often span over larger intervals.

Inspired by recent breakthroughs in sequence modeling in the field of Natural language Processing (NLP), we propose a natural extension of Visual Transformers (ViT)~\cite{dosovitskiy2021an} to videos. We view a video as analog to a text paragraph to be classified efficiently. To that end, we sample sentences (images) uniformly from it and divide those to words (patches).
In NLP, the Transformer model~\cite{vaswani2017attention} has proven superior to other sequence modeling techniques such as RNNs. The Transformer builds on a multi-head self attention layer, that learns global attention over the elements in the sequence. Similarly, our approach relies entirely on transformers, for both the spatial and the temporal dimensions.


We introduce an action classification model composed exclusively of self-attention layers operating in the spatial and temporal directions.  We name our model \textbf{STAM} (Space Time Attention Model).  The input sequence, in our case, is the sequence of image patches extracted from the individual frames and linearly projected onto a patch embedding space. First a spatial and then a temporal Transformer encoder models are applied on top of this embedding sequence to extract a video level representation or attention weighting of the frames. By leveraging this attention mechanism, we claim that the video sequence can be temporally subsampled by a larger factor than has previously been achieved, without degradation of the classification accuracy. \\ Figure~\ref{fig:runtime_accuracy} demonstrates the trade-off between the accuracy and runtime of top action recognition methods. While previous models are either accurate or efficient, models trained with our method offer a good combination of both. For example, STAM-32F achieves comparable accuracy to X3D-XL while being $\times 40$ faster.  

The motivation behind the proposed method is that applying global self-attention over a sequence of input frames is the key to reduce the number of required frames, by allowing information from individual frames to be propagated globally across the entire sequence. 3D convolutions, on the other hand, extract information locally over a small temporal (and spatial) scale, and therefore require frames to be sampled on a lower scale (i.e. dense sampling).  
In the NLP domain, where transformers are mostly being used, there is no issue of temporal continuity (and sampling density) since the words in a sentence are not temporally continuous as frames in the video are. Hence, our approach is a unique advantage of using Transformers on video data. 

Subsampling the input video substantially reduces the computational load during training and inference, and furthermore, has the additional benefit of lowering the cost of retrieving input data. Indeed, in several applications there is a cost associated with retrieving input data from storage, or across a communications network. In such bandwidth limited applications most action recognition methods are prohibitively expensive for deployment, and methods like ours that rely on significantly less input data to operate, posses a clear advantage. To give an idea of such a scenario, suppose that there is a cost incurred for every access to a video frame located in storage. For typical methods, $30*16=480$ frames are accessed in order to perform inference on a video. Compared to our method which requires $16$ frames for the same video, the cost reduction we achieve is $30$-fold, in addition to the reduction in run-time. 

\begin{figure*} [ht!]
  \centering
  \includegraphics[width=0.75\linewidth]{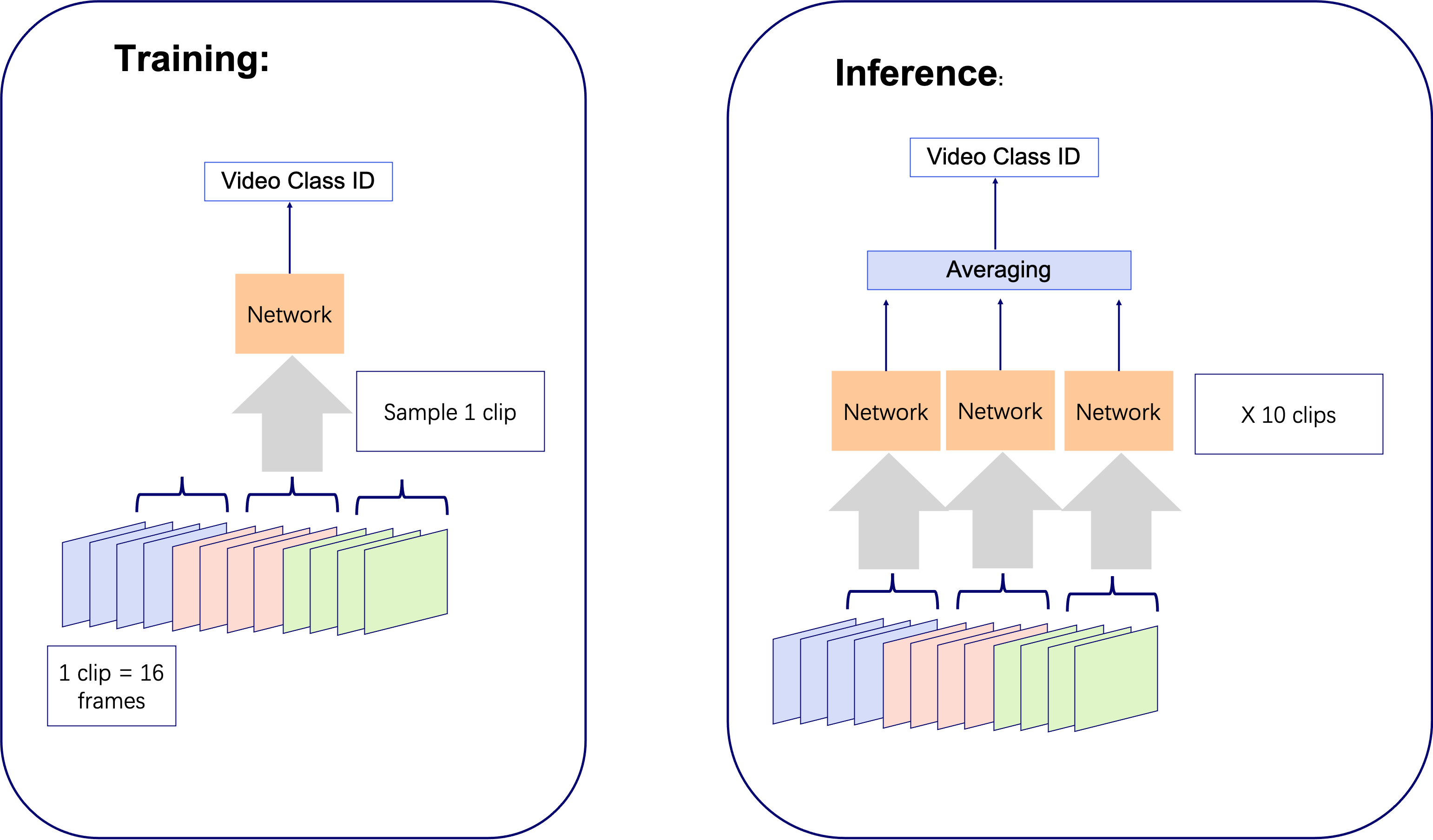}
  \caption{\textbf{Training and inference commonly used by other methods.} In the common approach multiple clips are sampled from each video, i.e., dense sampling of frames. Furthermore, training and inference have a different structure.}
  \label{fig:multi-clip-sampling}
\end{figure*}

An additional advantage of STAM is that it is an end-to-end trainable model.  This is both simpler to implement (requiring the same sampling strategy and model for train and inference) and has fewer hyperparameters.  Methods employing multi-clip averaging during test-time cannot be considered end-to-end trainable, since they add an additional temporal aggregation (averaging) layer during inference. Since it is used only at inference time, this additional layer is an ad-hoc component, whose effects are not taken into account during training. (See Figure~\ref{fig:multi-clip-sampling})

Our contributions can be summarized as follows:
\begin{itemize}
    \item We propose a novel method for video action recognition that is entirely based on transformers for representation of spatio-temporal visual data.
It is very simple, end-to-end trainable, and able to capture video information using only $3\%$ of the data processed by leading efficient methods.
\item Our method matches state-of-the-art accuracy while being more efficient by orders of magnitude. Specifically, on Kinetics-$400$ benchmark it achieves 78.8 top$1$-accuracy with $\mathbf{\times 40}$ faster inference time, or alternatively improving the efficient ECO~\cite{zolfaghari2018eco} accuracy by $\mathbf{+8\%}$ while being twice as fast. This makes it a leading solution for latency-sensitive video understanding. 
\end{itemize}








\section{Related Work}
\subsection{Transformers and Self Attention}
Self-attention, sometimes called intra-attention is an attention mechanism relating different positions of a single sequence in order to compute a representation of the sequence.
Self-attention layers are the building blocks of an encoder-decoder architecture called Transformer~\cite{vaswani2017attention}. 

The Transformer architecture has become the dominant model in the field of NLP, outperforming previous methods on  tasks, such as language translation, and text generation (~\cite{devlin2018bert},~\cite{vaswani2017attention}). Attempts to introduce Transformers to the computer vision domain~\cite{dosovitskiy2021an, touvron2020training} use only attention based layers instead of the commonly used convolutional layers, and produce state-of-the-art results on image classification benchmarks such as ImageNet.
Other methods (~\cite{carion2020end},~\cite{Hu_2018_CVPR}) combine convolutional networks with transformers for object detection.
Other methods that apply self-attention in vision tasks include ~\cite{Wang_2018_CVPR, pmlr-v80-parmar18a, NEURIPS2019_3416a75f, Cordonnier2020On}. These methods apply the Transformer model on the image pixel level, and therefore have to resort to approximations (either downsampling the image, or applying local attention instead of global).  
More similar to our approach, ~\cite{kalfaoglu2020late} applies a Transformer model in the domain of action recognition.  However, their model is a hybrid of a 3D convolutional model and a Transfomer model that acts on the CNN's output feature vectors.  Since it relies on 3D convolutions as part of the network, it has the same disadvantages of convolutional models (requiring dense sampling and a large number of frames), while our method is fully based on the Transformer model.


\subsection{Action Recognition}
Action recognition method typically operate by applying layers of 3D or 2D convolutions on spatio-temporal data~\cite{feichtenhofer2019slowfast, feichtenhofer2020x3d},~\cite{carreira2017quo, tran2018closer}.

X3D~\cite{feichtenhofer2020x3d} use network search and hyperparameter optimisation to find the network and sampling parameters (network depth, width, input spatial resolution, temporal resolution), while SlowFast~\cite{feichtenhofer2019slowfast} train two networks operating on different temporal resolutions. R3D~\cite{tran2018closer} decomposes the 3D convolution operator into two separate convolutions operating on the temporal and spatial dimensions. Although these works improve classification efficiency by modifying the network structure, they still require densely sampled frames as input. We overcome this limitation by removing the dependence of 3D convolutions, and modeling the temporal information via a self-attention sequence model. 

Several other works apply 2D convolutional networks on individual frames~\cite{lin2019tsm}, ~\cite{fan2019more}, and capture the temporal dependence by shifting feature maps in the temporal direction. In these methods information is propagated in a local neighborhood of frames, while in our case the global self-attention allows interaction across the whole spatial temporal dimensions. 



Another line of research is focused on reducing the computation cost of existing action recognition methods. These works introduce techniques to improve network efficiency by adaptive resolution sampling~\cite{meng2020ar}, importance clip sampling~\cite{korbar2019scsampler} or reducing redundant computation using linear approximations of feature maps~\cite{pan2021va}. However, these works still rely on multi-clip testing for inference, and thus suffer from the same type of inefficiency which our method proposes to solve.  We tackle the problem of computation by reducing the required input frames sampled from the video.

Additional works focus on action recognition with subsampled data. Mauthner et al.~\cite{mauthner2009action} suggested a method that uses a single-frame representation for short video sequences based on appearance and on motion information.
Other methods proposed encoding techniques for video representations~\cite{qiu2017deep,yue2015beyond}. Similarly to the methods that use video clips, they processed samples separately and fused the results over time.  
ECO~\cite{zolfaghari2018eco} sampled frames uniformly and applied a long-term spatio-temporal architecture over those. They learned per-frame representations with 2D CNNs and fed them into a 3D CNN afterwards. Our motivation and input data is similar, and the use of spatial and temporal encoders offers a significant improvement over their method.


\section{Method}

\begin{figure} [t!]
  \centering
  \includegraphics[width=\columnwidth]{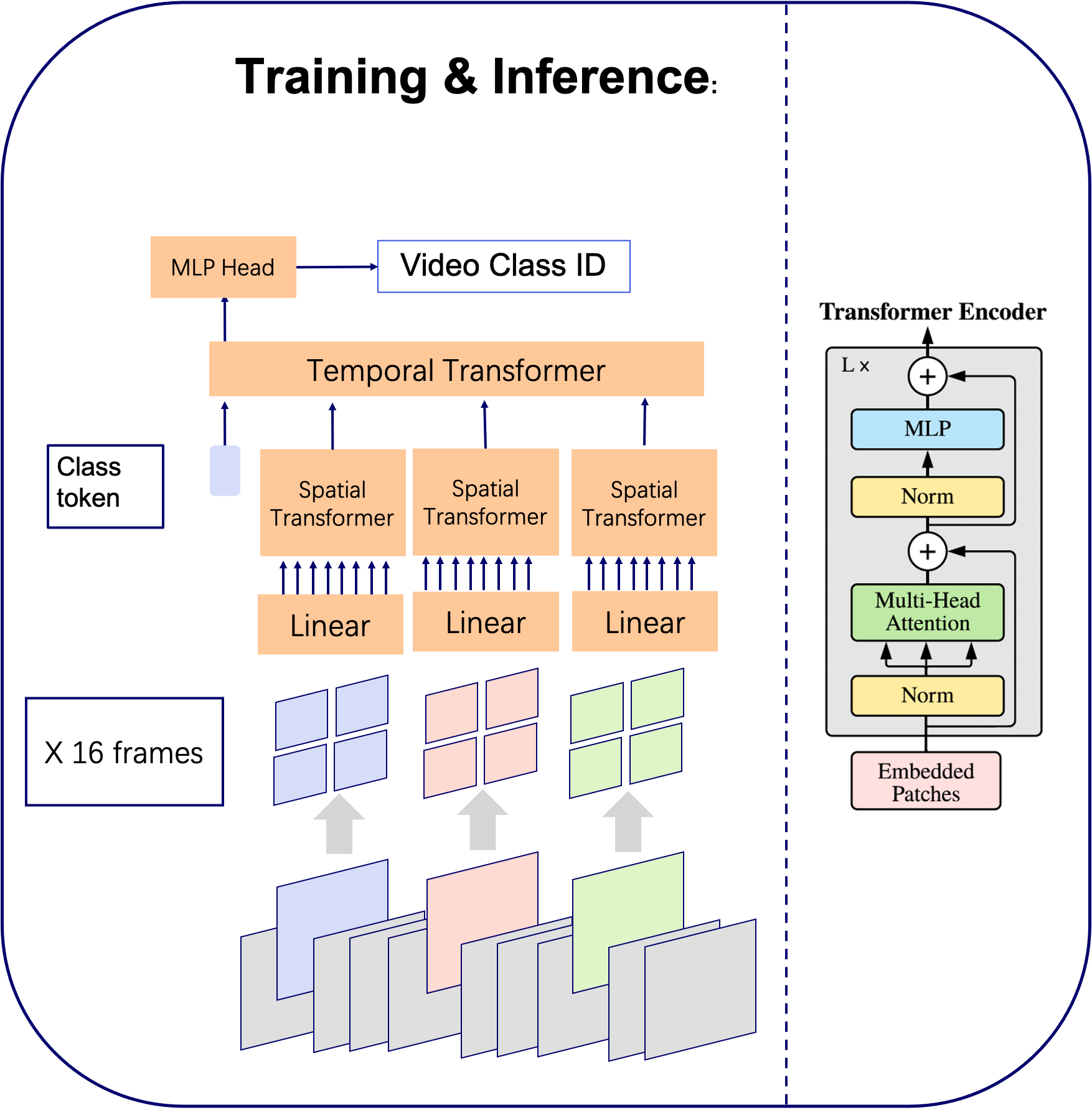}
  \caption{\textbf{Our proposed  Transformer Network for video}}
  \label{fig:overview}
\end{figure} 
\medskip
\medskip

\medskip
\medskip

As shown by the recent work on Visual Transformers (ViT)~\cite{dosovitskiy2021an}, self-attention models provide powerful representations for images, by viewing an image as a sequence of words, where each word embedding is equivalent to a linear projection of an image patch. 

In this section, We design a convolution-free model that is fully based on self-attention blocks for the spatio-temporal domain. By that, we offer an extension of ViT that captures temporal representations. While the proposed method is  focused on action recognition, we note that it can be easily modified for additional video understanding tasks.  

While the attention mechanism can be extended to temporal dependencies in several ways, 
our design is modular, making its implementation simpler and intuitive. 
Most importantly, our design naturally leverages the advantages of attention mechanisms compared to convolution operators when it comes to better utilization of temporal information. 

Our goal is to provide a model that can utilize sparsely subsampled temporal data for accurate predictions. Such model need to be able to capture long-term dependencies as well. 
While 2D convolutions filters are tailor-made for the structure of images, utilizing local connections and providing desired properties for object recognition and detection~\cite{kauderer2017quantifying}, the same properties might negatively affect the processing of subsampled temporal data. While a series of 3D-convolutions \textit{can} learn long-term interactions due to increased receptive field, they are biased towards local ones.  
In order to verify this, we conducted an experiment: we fed leading methods that are based on 3D convolutions with the same subsampled data as in our method. The results are presented in Table~\ref{Table:single_clip}. The performance of both methods degraded significantly, 
the error of X3D increased by $23\%$   and SlowFast error by $50\%$.

Transformers offer advantages over their convolutional counterparts regarding modeling long-term dependencies. While a multi-head self-attention
layer with sufficient number of heads is at least as expressive as any convolutional
layer~\cite{cordonnier2019relationship}, it also has the ability of
directly model long-distance interactions~\cite{ramachandran2019stand}.



We propose a combined \emph{spatial and temporal transformer} (STAM) which takes a sequence of frames sampled from the video as input, and outputs a video level classification prediction. As illustrated in Figure~\ref{fig:overview},
we process the sampled frames with a spatial transformer following the method of~\cite{dosovitskiy2021an}, and aggregate the resulting frame embeddings with a \emph{temporal transformer}.  In this way we separate the spatial attention (on each frame) from the temporal attention applied to the sequence of frame embedding vectors.  This separation between the spatial and temporal attention components has several advantages.  First, this reduces the computation by breaking down the input sequence into two shorter sequences.  In the first stage each patch is compared to $N$ other patches within a frames.  The second stage compares each frame embedding vector to $F$ other vectors, resulting in less overall computation than comparing each patch to $NF$ other patches.

The second advantage stems from the understanding that temporal information is better exploited on a higher (more abstract) level of the network.  In many works the where 2D and 3D convolutions are used in the same network, the 3D components are only used on the top layers.  Using the same reasoning, we apply the temporal attention on frame embeddings rather than on individual patches, since frame level representations provide more sense of what's going on in a video compared to individual patches.

\paragraph{Input embeddings.} The input to the spatio-temporal transformer is $X \in \mathbb{R}^{H \times W \times 3 \times F}$ consists of $F$ RGB frames of size $H \times W$ sampled from the original video. Each frame in this input block is first divided into non-overlapping patches. For a frame of size $H \times W$, we have $\smash{N=HW/P^2}$ patches of size $P \times P$. 

These patches are flattened into vectors and linearly projected into an embedding vector: 

\vspace{-0.2cm}
\begin{equation}
 {\bf z}_{(p,t)}^{(0)} = E {\bf x}_{(p,t)} + {\bf e}_{(p,t)}^{pos}
 \end{equation}
%
where input vector $\smash{{\bf x}_{(p,t)} \in \mathbb{R}^{3 P^2}}$, and embedding vector $\smash{{\bf z}_{(p,t)} \in \mathbb{R}^{D}}$ are related by a learnable positional embedding vector ${\bf e}_{(p,t)}^{pos}$, and matrix $E$.  The indices $p$, and $t$ are the patch and frame index, respecitvely with $p=1,\hdots,N$, and $t=1,\hdots,F$. In order to use the Transformer model for classification, a learnable classification token is added in the first position in the embedding sequence $\smash{{\bf z}_{(0,0)}^{(0)} \in      \mathbb{R}^{D}}$. As will be shown, this classification token will be used to encode the information from each frame and propagate it temporally across the sequence of frames. For this reason we include a separate classification token for each frame in the sequence $\smash{{\bf z}_{(0,t)}^{(0)}}$.

\paragraph{Multi-head Self-Attention block (MSA).} 

STAM consists of $L$ MSA blocks. At each block $\ell \in \{1,\hdots,L\}$, and head $a \in \{1,\hdots,\mathcal{A}\}$, each patch representation is transformed into query, key, and value vectors. The representation produced by the previous block ${\bf z}_{(p,t)}^{\ell-1}$ is used as input. 
\begin{align}
{\bf q}_{(p,t)}^{(\ell,a)} &= W_Q^{(\ell,a)} \mathrm{LN}\left({\bf z}_{(p,t)}^{(\ell-1)}\right) \in \mathbb{R}^{D_h}\\
{\bf k}_{(p,t)}^{(\ell,a)} &= W_K^{(\ell,a)} \mathrm{LN}\left({\bf z}_{(p,t)}^{(\ell-1)}\right) \in \mathbb{R}^{D_h}\\
{\bf v}_{(p,t)}^{(\ell,a)} &= W_V^{(\ell,a)} \mathrm{LN}\left({\bf z}_{(p,t)}^{(\ell-1)}\right) \in \mathbb{R}^{D_h}
\end{align}
Where $\mathrm{LN}$ represents a LayerNorm~\cite{ba2016layer}. The dimension of each attention head is given by $\smash{D_h = D / \mathcal{A}}$. 

The attention weights are computed by a dot product comparison between queries and keys.  The self-attention weights $\smash{\pmb{\alpha}_{(p,t)}^{(\ell,a)} \in \mathbb{R}^{NF+F}}$ for patch $(p,t)$ are given by:

\vspace{-0.3cm}
\begin{align}
\pmb{\alpha}_{(p,t)}^{(\ell,a)} &= \mathrm{SM}\left( \frac{{\bf q}_{(p,t)}^{(\ell,a)}}{\sqrt{D_h}}^\top \cdot \left[{\bf k}_{(0,t)}^{(\ell,a)} \left\{ {\bf k}_{(p',t')}^{(\ell,a)} \right\}_{\begin{subarray}{l}p'=1,...,N\\t'=1,...,F
\end{subarray}}  \right] \right)
\label{eq:attentionST}
\end{align}
%
where $\mathrm{SM}()$ denotes the softmax activation function, and ${\bf k}_{(0,t)}^{(\ell,a)}$ is the key value associated with the class token of frame $t$. 

\paragraph{Spatial attention} Applying attention over all the patches of the sequence is computationally expensive, therefore, an alternative configuration is required in order to make the spatio-temporal attention computationally tractable. 
A reduction in computation can be achieved by disentangling the spatial and temporal dimensions. For the spatial attention, we apply attention between patches of the same frame:

\vspace{-0.3cm}
\begin{align}
\pmb{\alpha}_{(p,t)}^{(\ell,a)\mathrm{space}} &= \mathrm{SM}\left( \frac{{\bf q}_{(p,t)}^{(\ell,a)}}{\sqrt{D_h}}^\top \cdot \left[{\bf k}_{(0,t)}^{(\ell,a)}, \left\{ {\bf k}_{(p',t)}^{(\ell,a)} \right\}_{p'=1,...,N}  \right] \right).\label{eq:attentionSPACE}
\end{align}
\vspace{-0.3cm}

The attention vector entries are used as coefficients in a weighted sum over the values for each attention head:

\vspace{-0.3cm}
\begin{align}
{\bf s}_{(p,t)}^{(\ell,a)} &= {\alpha}_{(p,t),(0,t)}^{(\ell,a)} {\bf v}_{(0,t)}^{(\ell,a)} + \sum_{p'=1}^N {\alpha}_{(p,t),(p',t)}^{(\ell,a)} {\bf v}_{(p',t)}^{(\ell,a)}.\label{eq:attn_prod}
\end{align}

These outputs from attention heads are concatenated and passed through a 2 Multi-Layer Perceptron (MLP) layers with GeLU~\cite{Hendrycks2016GaussianEL} activations:

\vspace{-0.3cm}
\begin{align}
{\bf z'}_{(p,t)}^{(\ell)} &= W_O \left[ \begin{array}{c} {\bf s}_{(p,t)}^{(\ell,1)}\\ \vdots \\ {\bf s}_{(p,t)}^{(\ell,\mathcal{A})} \end{array} \right] + {\bf z}_{(p,t)}^{(\ell-1)} \label{eq:hiddenstate}\\
{\bf z}_{(p,t)}^{(\ell)} &= \mathrm{MLP}\left(\mathrm{LN}\left({\bf z'}_{(p,t)}^{(\ell)}\right)\right) + {\bf z'}_{(p,t)}^{(\ell)}.\label{eq:encoding}
\end{align}
The MSA and MLP layers are operating as residual operators thanks to added skip-connections

After passing through the spatial Transformer layers, the class embedding from each frame is used to produce an embedding vector ${\bf f}_t$.  This frame embedding will be fed into the temporal attention.
\vspace{-0.3cm}
\begin{equation}
{\bf f}_t = \mathrm{LN}\left({\bf z}_{(0,t)}^{(L_{space})}\right)_{t=1,...,F}  \in  \mathbb{R}^{D}.
\label{eq:frame_emb}
\end{equation} 
%
where $L_{space}$ is the number of layers of the spatial Transformer.

\paragraph{Temporal attention.} The spatial attention provides a powerfull representation for each individual frame by applying attention between patches in the same image.  However, in order to capture the temporal information across the frame sequence, a temporal attention mechanism is required.  The effect of temporal modeling can be seen in table~\ref{Table:temporal_aggregator}.  The spatial attention backbone provides a good representation of the videos, however the additional temporal attention provides a significant improvement over it.
In our model, the temporal attention layers are applied on the representations produced by the spatial attention layers. 

For the temporal blocks of our model, we use the frame embedding vectors from eqn.~\ref{eq:frame_emb}, stacked into a matrix $X_{time} \in \mathbb{R}^{F\times D}$ as the input sequence. As before, we add a trainable classification token at index $t=0$. This input sequence is projected into query/key/value vectors ${\bf q}^{(\ell,a)}_t$, ${\bf k}^{(\ell,a)}_t$, ${\bf v}^{(\ell,a)}_t$. The temporal attention is then computed only over the frame index.

\vspace{-0.3cm}
\begin{align}
\pmb{\alpha}_{t}^{(\ell,a)\mathrm{time}} &= \mathrm{SM}\left( \frac{{\bf q}_{t}^{(\ell,a)}}{\sqrt{D_h}}^\top \cdot \left[{\bf k}_{0}^{(\ell,a)} \left\{ {\bf k}_{t'}^{(\ell,a)} \right\}_{t'=1,...,F}  \right] \right).
\end{align}
\vspace{-0.3cm}

Next, we apply the same equations of the attention block eqn.~\eqref{eq:attn_prod} through~\eqref{eq:encoding}, with a single axis describing the frame indices instead of the double $(p,t)$ index which was used in those equations. The embedding vector for a video sequence is given by applying the layer norm on the classification embedding from the top layer: 
\vspace{-0.3cm}
\begin{equation}
{\bf y} = \mathrm{LN}\left({\bf z}_{(0)}^{(L_{time})}\right) \in  \mathbb{R}^{D}.
\end{equation}
%
where $L_{time}$ is the number of temporal attention layers.
An additional single layer MLP is applied as the classifier, outputing a vector of dimension equal to the number of classes.

The added cost of the temporal encoder layers over the spatial layers is negligible since they operate on an input sequence of length $F$, which is an order of magnitude smaller than the number of patches $N$ (in our experiments usually $F\!=\!16$, while $N\!=\!196$).  In this architecture, the total complexity is $O(FN^2\!+\!F^2)$.  If the attention operation were applied over all spatio-temporal patches, the complexity would be in the order of $O((FN)^2)$, which is prohibitively large.

An alternative possible configuration for the spatio-temporal attention would be to apply the temporal attention (between patches at the same spatial position, but from different frames), after the spatial attention within each block. We found that this variation produces slightly worse results, and higher computational cost. 

Using an analogy to NLP, we consider each frame to be a sentence (in which patches play the role of word tokens), and each video is a paragraph of sentences. Following this analogy, it makes sense to apply a transformer separately on the sentences, extracting a representation vector per sentence, and then an additional Transformer on these vectors to predict a class (e.g. sentiment) from the entire paragraph.


\begin{table*}[ht!]
\centering
\begin{tabular}{|l|c|c|c|c|c|} 
\hline
\multicolumn{1}{|c|}{Model}  & \begin{tabular}[c]{@{}c@{}}Top-1 Accuracy\\{[}\%] \end{tabular} & \begin{tabular}[c]{@{}c@{}}Flops $\times$ views\\{[}G] \end{tabular} & \begin{tabular}[c]{@{}c@{}}Param\\{[}M] \end{tabular} & \begin{tabular}[c]{@{}c@{}}Runtime\\{[}hrs] \end{tabular} & \begin{tabular}[c]{@{}c@{}}Runtime\\{[}VPS] \end{tabular}\\ 
\hline
Oct-I3D + NL~\cite{chen2019drop}                                                                                                     & $75.7$                                                         & $28.9 \times 30$            &$33.6$      & ---  & ---      \\
SlowFast $8\times8$, R50                                                                                                         & $77.0$                                                         & $65.7 \times 30$                      &$34.4$              & $2.7$5 & $1.4$ \\
X3D-M                                                                                                              & $76.0$                                                        & $\mathbf{6.2 \times 30}$                                   &$3.8$            & $1.47$   & $1.3$ \\
X3D-L                                                                                                             & $77.5$                                                          & $24.8 \times 30$                                       & $6.1$             & $2.06$   & $0.46$ \\
X3D-XL                                                                                                             & $\mathbf{79.1}$                                                          & $48.4 \times 30$                                       & $11.0$             & ---    & --- \\
TSM R50                                                                  & $74.7$                                  & $65 \times 10$                                         & $24.3$             &  ---    & --- \\
Nonlocal R101                                                            & $77.7$                                  & $359 \times 30$                                        & $54.3$              &  ---    & --- \\
\hline
\textbf{STAM ($\mathbf{16}$ Frames)}                                                & $\mathbf{79.3}$                         &$\mathbf{270 \times 1}$                                         & $96$         & $\mathbf{0.05}$    &$\mathbf{20.0}$ \\
\textbf{STAM ($\mathbf{64}$ Frames)}                                                & $\mathbf{80.5}$                         &$1080 \times 1$                                        & $96$         & $\mathbf{0.21}$  & $\mathbf{4.8}$   \\
\hline
\end{tabular}
\medskip
\caption{\textbf{Model comparison on Kinetics400}. Time measurements were done on Nvidia V100 GPUs with mixed precision. The Runtime [hrs] measures inference on Kinetics-400 validation set (using 8 GPUs), while the videos per second (VPS) measurement was done on a single GPU. Results of various methods are as reported in the relevant publications. The proposed STAM is an order of magnitude faster while providing SOTA accuracy.}
    \label{Table:models_comparison}
\end{table*}

\section{Experiments}

\paragraph{Implementation Details} 
STAM consists of two parts: the spatial attention, and the temporal attention. In our experiments we closely follow the ViT\_B model proposed by~\cite{vaswani2017attention} as the spatial Transformer.  This model is the lighter verion of the ViT family of models and contains 12 MSA layers, each with 12 self-attention heads.  We use the imagenet-21K pretraining provided by~\cite{vaswani2017attention} (unless specified otherwise).  For the \emph{Temporal Transformer} we use an even smaller version of the Transformer with 6-layers and 8-head self-attention ($L_{space}=12$, $L_{time}=6$).  The temporal layers are trained from scratch, and initialized randomly. 

We sample frames uniformly across the video. For training we resize the smaller dimension of each frame to a value $\in [256,320]$, and take a random crop of size $224 \times 224$ from the same location for all frames of the same video. We also apply random flip augmentation, Cutout with factor $0.5$, and auto-augment with Imagenet policy on all frames.

For inference we resize each frame so that the smaller dimension is 256, and take a crop of size $224\times 224$ from the center. We use the same uniform frame sampling for training and inference.

\begin{table*}[ht!]
\centering
\begin{tabular}{|l|c|c|} 
\hline
\multicolumn{1}{|c|}{Backbone+Temporal} &   \begin{tabular}[c]{@{}c@{}}Top-1 Accuracy\\{[}\%] \end{tabular} & \begin{tabular}[c]{@{}c@{}}Flops\\{[}G] \end{tabular}  \\ 
\hline
ViT+Temporal-Transformer              & \textbf{77.8}                                                                                 & 270     \\
TResNet-M+Temporal-Transformer              & 75.7                                                                                 & 93      \\
ViT+Mean                & 75.1                                               & 265         \\
TResNet-M+Mean                & 71.9                                               & 88         \\
\hline
\end{tabular}
\medskip
\caption{\textbf{Temporal Transformer vs. Mean.} Comparing the Temporal Transformer representation vs. mean of frame embeddings.}
    \label{Table:temporal_aggregator}
\end{table*}

\begin{table*}[ht!]
\centering
\begin{tabular}{|l|c|c|} 
\hline
\multicolumn{1}{|c|}{Model (num. of frames)} &   \begin{tabular}[c]{@{}c@{}}Top-1 Accuracy\\{[}\%] \end{tabular} & \begin{tabular}[c]{@{}c@{}}Flops\\{[}G] \end{tabular}  \\ 
\hline
TResNet-M+Temporal (16)             & 75.7                                                                                 & 93     \\
TResNet-L+Temporal (8)             & 75.9                                                                                 & 77     \\
ResNet50+Temporal (16)             & 72.5                                                                                 & 70      \\
ViT-B+Temporal (16)                & \textbf{79.3}                                               & 270         \\
\hline
\end{tabular}
\medskip
\caption{\textbf{Using different backbones with the Temporal Transformer}. TResNet and ViT models use imagenet-21K pretraining, while ResNet50 is used with imagenet-1K pretraining.}
    \label{Table:backbone_comparison}
\end{table*}

\paragraph{Training} For Kinetics-400 we train our model on 8V100 GPUs for 30 epochs with learning rate warm-up, and a cosine learning rate schedule.
Compared to X3D and SlowFast, both trained with 128 GPUs for 256 epochs, our training is much faster and requires less epochs ($\sim 30$). 

\paragraph{Kinetics:} We compare our method to others on Kinetics-400 dataset~\cite{kinetics}. Table~\ref{Table:models_comparison} shows our method achieves 79.3\% top-1 accuracy using 16 sampled frames per video (at 270 GFLOPS). Compared to X3D\_L, which achieves similar top-1 accuracy (77.5\%) using 30 clips for inference (at 744 GFLOPS), this is an 1.8 improvement in top-1 accuracy, using only $~36\%$ of the computation required by X3D\_L.  

The reduction in run-time is even more significant.  Compared to X3D\_L we observe a reduction in inference time from 2.27 to 0.05 hrs for the entire validation set. We also calculate the video per second runtime by performing inference on a single batch of clips.  We find that our method (with 16 frames) is able to outperform X3D\_L by a factor of $\bf 43$. This substantial improvement in runtime is partly due to the fewer input frames required by our method, and in part due to the improved efficiency of the ViT\_B model compared to the more complex X3D architecture.  Runtime is a more tangible metric for efficiency, and therefore we focus on it.

\begin{table*}[ht!]
\centering
\begin{tabular}{|c|c|c|} 
\hline
\multicolumn{1}{|c|}{Number of frames} &   \begin{tabular}[c]{@{}c@{}}Top-1 Accuracy\\{[}\%] \end{tabular} & \begin{tabular}[c]{@{}c@{}}Flops\\{[}G] \end{tabular}  \\ 
\hline

4              & 74.1                                                                                 & 67      \\
8              & 77.1                                                                                  &135     \\
16                & \textbf{79.3}                                               & 270         \\
32                & \textbf{79.9}                                               & 540         \\
64            &\textbf{80.5}                                                    & 1080        \\ 
\hline
\end{tabular}
\medskip
\caption{\textbf{Number of frames used for prediction} Comparing different number of frames sampled uniformly from the video as input, using the ViT+Temporal model.}
    \label{Table:sequence_length}

\end{table*}


\begin{table*}[ht!]
\centering
\begin{tabular}{|l|c|} 
\hline
\multicolumn{1}{|c|}{Method} &   \begin{tabular}[c]{@{}c@{}}Top-1 Accuracy\\{[}\%] \end{tabular}  \\ 
\hline
X3D\_L              & 72.25 (77.5)                         \\
SlowFast\_8x8\_R50                & 65.5 (77.0)          \\
\hline
\end{tabular}
\medskip
\caption{\textbf{Evaluating X3D and SlowFast with one clip (16 frames)}. Applying the same sampling strategy that we use in Video Transformer on other methods. (In parentheses - accuracy at 30 views per video).}
    \label{Table:single_clip}
\end{table*}

\subsection{Ablation Experiments}
This section provides ablation studies on Kinetics-400
comparing accuracy and computational complexity.

First, we compare models with only spatial attention to a model that has temporal attention as well. In table~\ref{Table:temporal_aggregator} we compare these two types of models and verify the positive effect of the temporal attention.  We compare two variants of our method. The first is our full method using ViT\_B backbone as the spatial attention model  with a temporal transformer, and the second uses the same backbone but replace the temporal transformer with an average of the frame embeddings. The table shows that the temporal transformer provides a gain of 2.7 over the naive approach. Additionally, we repeat this experiment using a different backbone and see the consistency of the result across different backbones.  We use a TResNet-M~\cite{Ridnik_2021_WACV} backbone, and again observe that the temporal transformer significantly improves accuracy.

In table~\ref{Table:backbone_comparison}, we show the effect of replacing the ViT model with a CNN model. We compare 3 different CNN variants (TResNet-M, TResNet-L~\cite{Ridnik_2021_WACV}, ResNet50) trained together with the temporal transformer. This experiment shows that the highest accuracy is achieved using the full spatial, and temporal transformer model. Although replacing the spatial transformer with a CNN model is possible, and achieves reasonable accuracy, it is less powerful than the combination of the spatial and temporal attention, which applies attention over all the patches of the frame sequence.

\paragraph{Number of frames.} In table~\ref{Table:sequence_length} we compare different sequence lengths as input to STAM. We sample 8, 16, 32, and 64 frames and compare the results.  The clear trend is an increase in accuracy along with the increase in sequence length.  For an increase of 16 additional frames (from 16 frames to 32 frames) we see an improvement of 0.6\% to the accuracy.  Switching the number of input frames from 32 to 64, results in a similar gain of $0.6\%$. Using a larger number of frames doesn't improve the accuracy. 

The use of 4 frames to classify a video of length 10 seconds suggests the use of our method for longer range actions.  For instance, if we use the same sampling rate for a 1 minute video, we would require 24 frames for inference.

\paragraph{Runtime comparison.}  In table~\ref{Table:single_clip} we evaluate the accuracy of two leading methods, with a reduced number of input frames.  We use a single 16 frame input clip sampled uniformly, and so use these methods with the a similar running time to our method.  The table clearly shows that by reducing the number of input frames, these methods suffer a large degradation in accuracy.  This show that methods that rely on 3D convolutions require frames to be sampled at a higher rate than the one we use, and cannot be made to improve their runtime by sampling sparse input frames.  

Figure~\ref{fig:runtime_accuracy} plots different action recognition methods on the accuracy vs. runtime (VPS) scale.  We compare methods that are both designed for efficiency (ECO~\cite{zolfaghari2018eco}), and methods that apply heavier models for increased accuracy (X3D~\cite{feichtenhofer2020x3d}, SlowFast~\cite{feichtenhofer2019slowfast}).  We see that our method is on par with the accuracy of the slower methods and improves over their runtime by a significant margin.  STAM's runtime is comparable to that of ECO, at 20 VPS, yet significantly outperforms that method in terms of accuracy by $8\%$.  Since ECO employs a similar sampling strategy to ours (sampling individual frames across the video), we conclude that the temporal transformer is better at capturing the temporal information from separate frames.



\section{Conclusion}
In this work we have presented a method for efficient video action recognition that is entirely based on transformers. Inspired by NLP, we model a video as a paragraph and uniformly select frames that are modeled as sentences. This modeling allows us to utilize transformers to capture complex spatio-temporal dependencies between distinct frames, leading to accurate predictions based on a small fraction of the video data. 
The accuracy of our models' predictions is comparable to state-of-the-art methods while being faster by orders of magnitude, making it a good candidate for latency-sensitive applications like real-time, such as recognition and video retrieval.
In addition, our method is simple, end-to-end and generic, thus, it can be used for further video understanding tasks.


{\small
\bibliographystyle{ieee_fullname}
\bibliography{references}
}

\clearpage
\appendix
\newpage

\begin{center} \begin{huge}{Appendix} \end{huge} \end{center}
\label{appendix}

\section{Additional Datasests}

We evaluate our method on 2 additional datasets: UCF101, Charades. 




\paragraph{UCF101}

UCF101~\cite{soomro2012ucf101} is an older, yet still popular action recognition dataset.  Our results on this dataset are presented in table~\ref{Table:ucf101}. The results suggest that our model achieves a good trade-off between runtime and accuracy. 

\begin{table}[ht!]
\centering

\begin{tabular}{|l|c|c|} 
\hline
\multicolumn{1}{|c|}{Model}  & \begin{tabular}[c]{@{}c@{}}Top-1 Accuracy\\{[}\%] \end{tabular}  & \begin{tabular}[c]{@{}c@{}}Runtime\\{[}VPS] \end{tabular}\\ 
\hline
ECO ~\cite{zolfaghari2018eco}                                                                 & 93.6                           &   \textbf{20.8}     \\
R(2+1)D-34                                                                  & 96.8
            & N/A  \\
I3D                                                            &  95.6                                 & N/A            \\
S3D                                                            &  96.8                            & N/A                 \\
FASTER32~\cite{zhu2020faster}                                                            &  96.9                        & 2.8                      \\
\hline
\textbf{STAM-32}                                                & \textbf{97.0}                     & 10     \\
\hline
\end{tabular}
\medskip
\caption{\textbf{Results on UCF-101 dataset}.  Results of various methods are as reported in the relevant publications. We compare methods that use only RGB frames as input (without Optical Flow), and are pretrained on Kinetics-400 or Imagenet }

    \label{Table:ucf101}
\end{table}

\paragraph{Charades}
Charades~\cite{sigurdsson2016hollywood} is a dataset with longer range interactions, and multiple labels per video. Table~\ref{Table:charades} presents our results on this dataset.  Although our model doesn't reach SotA accuracy, it shows promise as an efficient model, requiring less input frames from each video, and less FLOPS. 

\begin{table}[ht!]
\centering
\begin{tabular}{|l|c|c|} 
\hline
\multicolumn{1}{|c|}{Model}  & \begin{tabular}[c]{@{}c@{}}Top-1 Accuracy\\{[}\%] \end{tabular} & \begin{tabular}[c]{@{}c@{}}Flops $\times$ views\\{[}G] \end{tabular} \\ 
\hline
Nonlocal                                                                                                         & 37.5                 & 544 $\times$ 30                                                  \\
STRG, +NL                                                                                                         & 39.7                    & 630 $\times$ 30                                      \\
Timeception                                                                                                         & 41.1                      & N/A                                   \\
LFB, +NL                    & 42.5                      & 529 $\times$ 30                         \\
SlowFast, +NL                   & 42.5              & 234 $\times$ 30                                   \\
X3D-XL                          & $ \textbf{43.4}$              & 48 $\times$ 30                                       \\
\hline
STAM-64                     & 39.7                      & $\mathbf{1040 \times 1}$                                    \\
\hline
\end{tabular}
\medskip
\caption{\textbf{Results on Charades dataset}.  Results of various methods are as reported in the relevant publications. Including methods that are pretrained on Kinetics-400 or Imagenet.}
    \label{Table:charades}
\end{table}


\end{document}